\pdfoutput=1

\documentclass[11pt]{article}

\usepackage{EMNLP2022}

\usepackage{times}
\usepackage{latexsym}

\usepackage[T1]{fontenc}

\usepackage[utf8]{inputenc}

\usepackage{microtype}

\usepackage{inconsolata}

\usepackage{graphicx} 
\usepackage{booktabs} 
\usepackage{multirow} 
\usepackage{amsmath} 
\usepackage{amsfonts} 
\usepackage[figuresright]{rotating}

\title{Mixture of Attention Heads: 
Selecting Attention Heads Per Token}

 \author{Xiaofeng Zhang$^{1,2}$\thanks{~~Equal contribution. \texttt{xiaofeng\_z@buaa.edu.cn}, \texttt{yikang.shn@gmail.com}}, Yikang Shen$^{3,4}$\footnotemark[1], Zeyu Huang$^{1,2}$, Jie Zhou$^4$, Wenge Rong$^1$, Zhang Xiong$^1$ \\
        $^1$ State Key Laboratory of Software Development Environment, \\ School of Computer Science and Engineering, Beihang University, China \\
        $^2$ Sino-French Engineer School, Beihang University, China \\ 
        $^3$ Mila, University of Montreal, Canada \\
        $^4$ Wechat AI, Tencent, China
        }

\begin{document}
\maketitle
\begin{abstract}
Mixture-of-Experts (MoE) networks have been proposed as an efficient way to scale up model capacity and implement conditional computing.
However, the study of MoE components mostly focused on the feedforward layer in Transformer architecture.
This paper proposes the Mixture of Attention Heads (MoA), a new architecture that combines multi-head attention with the MoE mechanism.
MoA includes a set of attention heads that each has its own set of parameters. 
Given an input, a router dynamically selects a subset of $k$ attention heads per token.
This conditional computation schema allows MoA to achieve stronger performance than the standard multi-head attention layer.
Furthermore, the sparsely gated MoA can easily scale up the number of attention heads and the number of parameters while preserving computational efficiency.
In addition to the performance improvements, MoA also automatically differentiates heads' utilities, providing a new perspective to discuss the model's interpretability.
We conducted experiments on several important tasks, including Machine Translation and Masked Language Modeling.
Experiments have shown promising results on several tasks against strong baselines that involve large and very deep models\footnote{The code can be found at \url{https://github.com/yikangshen/MoA}.}.
\end{abstract}

\section{Introduction}

\begin{figure}[t]
    \centering
    \includegraphics[width=0.6\columnwidth]{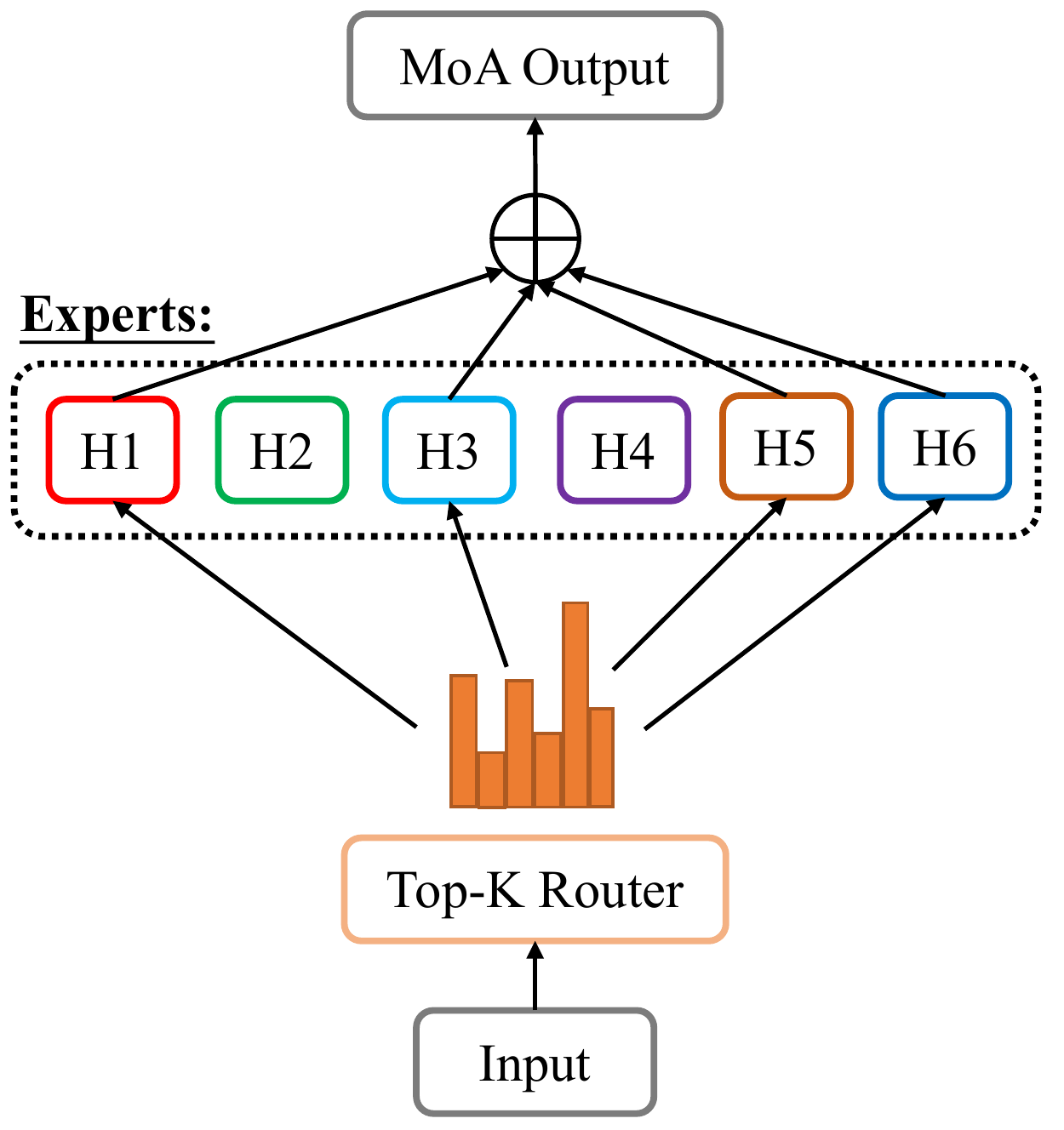}
    \caption{Simple illustration of MoA. MoA consists of a set of attention heads named attention experts. For each token in the input, a Router selects $k$ attention heads among all attention experts with different confidences. The output is a weighted sum of the selected attention heads given the confidence calculated by the Router.}
    \label{fig:MoAillus}
\end{figure}

In recent years, large models have become a popular trend in the research of Natural Language Processing, especially large-scale Transformer~\citep{DBLP:conf/nips/VaswaniSPUJGKP17}. 
The model's capacity has increased from millions of parameters~\citep{DBLP:conf/naacl/DevlinCLT19,DBLP:journals/corr/abs-1907-11692}, to billions of parameters~\citep{DBLP:journals/corr/abs-1909-08053,DBLP:journals/jmlr/RaffelSRLNMZLL20,DBLP:journals/corr/abs-2203-00555}, even to trillions of parameters~\citep{DBLP:journals/corr/abs-2112-06905,DBLP:journals/corr/abs-2101-03961}. However, these large-scale models demand substantially more computations than small-scale models. A popular trend is to utilize conditional computation with a sparsely activated model to seek greater computational efficiency. Thus, only a part of the model’s parameters is used for a specific input during the forward computation, which alleviates the computational load. 

Among these attempts, the Mixture of Experts (MoE)~\citep{DBLP:journals/neco/JacobsJNH91,DBLP:journals/neco/JordanJ94} is an essential technique. 
Since first applying the mixture of experts to Transformer architecture~\citep{DBLP:conf/nips/ShazeerCPTVKHLH18}, researchers have mainly focused on combining the Feed-Forward Network layer and the Mixture of Experts. 
Recent works have discussed how to get a better routing strategy~\citep{DBLP:conf/iclr/ShazeerMMDLHD17,DBLP:journals/corr/abs-2110-08246,DBLP:conf/icml/LewisBDGZ21,DBLP:journals/corr/abs-2112-14397} or how to scale up the Mixture of Experts on different nodes of GPUs~\citep{DBLP:conf/iclr/LepikhinLXCFHKS21,DBLP:journals/corr/abs-2101-03961}. 
However, few attempts have explored the possibility of combining MoE with the Multi-Head Attention (MHA) mechanism. 
Since the MHA is another compulsory module in the Transformer architecture, combining MoE and the attention mechanism could also help achieve better performance while restraining the computational cost.

Besides, previous research has investigated the utility of different attention heads. \citet{DBLP:conf/acl/PengSLS20} found that the combination (reallocation) of a subset of attention heads helps the Translation task since they prune the useless attention heads. In the field of dependency parsing, researchers have unveiled that some attention heads in BERT-like language models~\citep{DBLP:conf/naacl/DevlinCLT19,DBLP:journals/corr/abs-1907-11692} model individual dependency types~\citep{DBLP:journals/corr/abs-1911-12246} and syntactic functions~\citep{shen2020unsupervised}. ~\citet{DBLP:conf/acl/VoitaTMST19} claimed that the attention heads have different functions that could be categorized into three types.
There is no need to pass through all multiple attention heads for an input token if we could select some relevant attention heads whose function is proper.
Thus, we conceive an attention mechanism that selects different attention heads per token.


Based on the above discussion, we proposed Mixture of Attention Heads (MoA) (Section~\ref{sec:moa}), an attention mechanism that selects different attention heads for different inputs. A simple illustration of this idea is shown in Figure~\ref{fig:MoAillus}. MoA includes a set of of attention heads with different parameters. Given an input, a routing network dynamically selects a subset of $k$ attention heads for each token. The output is a weighted sum of the selected attention heads given the confidence calculated by the routing network.


We conducted experiments on two tasks: Machine Translation and Masked Language Modeling (Section~\ref{sec:results}). Experiments shown promising results against several strong baselines. In all tasks, our proposed mixture of attention heads outperforms the original Transformer architecture~\citep{DBLP:conf/nips/VaswaniSPUJGKP17}. Our model surpasses many large models or achieves comparable results with only a half computational cost. Our contributions can be summarized in three folds: 
1) We proposed a new attention mechanism called Mixture of Attention Heads, combining the idea of Mixture of Experts with the attention mechanism. 
2) MoA can improve the model's performance without substantially adding parameters and computational cost. 
3) MoA is easy to scale up while maintaining with a restrained computation complexity, resulting in a further performance amelioration.

\section{Related Work}
\paragraph{Mixture of Experts}
The Mixture of Experts (MoE) was firstly introduced in the 1990s~\citep{DBLP:journals/neco/JacobsJNH91,DBLP:journals/neco/JordanJ94}.
\citet{DBLP:conf/iclr/ShazeerMMDLHD17} adopted this method into modern deep learning architectures (LSTM;~\citealt{DBLP:journals/neco/HochreiterS97}) and proved its effectiveness in Language Modeling and Machine Translation. The MoE was used to substitute the FFN layers in Transformer architecture~\citep{DBLP:conf/nips/VaswaniSPUJGKP17} by the Mesh Tensorflow library~\citep{DBLP:conf/nips/ShazeerCPTVKHLH18}. Gshard~\citep{DBLP:conf/iclr/LepikhinLXCFHKS21} is a lightweight module that helps scale up multilingual neural machine translation Transformer with a Sparsely-Gated Mixture of Experts beyond 600 billion parameters. In Switch Transformer~\citep{DBLP:journals/corr/abs-2101-03961}, the authors scaled the MoE-integrated Transformer architecture toward trillion parameter models. GLaM~\citep{DBLP:journals/corr/abs-2112-06905} utilized a decoder-only architecture to do language model pre-training. \citet{DBLP:journals/corr/abs-2201-05596} proposed a Pyramid-Residual-MoE for smaller model size and fast inference.

Various routing strategies~\citep{DBLP:conf/iclr/ShazeerMMDLHD17,DBLP:journals/corr/abs-2110-08246,DBLP:conf/icml/LewisBDGZ21,DBLP:journals/corr/abs-2112-14397} have been investigated for stabilizing the MoE training and balancing the expert loads. \citet{DBLP:journals/corr/abs-2204-09179} pointed out the representation collapse issue in the sparse Mixture of Experts models and solved by a two-stage routing strategy.

\paragraph{Machine Translation Architectures}
With original Transformer architecture~\citep{DBLP:conf/nips/VaswaniSPUJGKP17}, ~\citet{ott-etal-2018-scaling} found that training with reduced precision and large batch could improve the translation performance.
Some models get better performance on translation by using larger scale of Transformer.
\citet{DBLP:conf/emnlp/LiuLGCH20} deepened the encoder and decoder of the Transformer by adequately initializing the model.
DeepNet~\citep{DBLP:journals/corr/abs-2203-00555} scaled Transformers up to 1,000 layers by introducing a new normalization function. However, these methods require a great amount of computational cost.
Some models make changes to the self-attention module. \citet{DBLP:conf/acl/PengSLS20} proposed MAE model. The reallocation of attention heads got better performance on Translation, since the model prune useless multi-head attention heads. However, their method is difficult to scale up and get further improvement of the results because it needs to use all the attention heads in the model rather than sparsely activate them. It also requires the complicated block coordinate descent training steps. ~\citet{wu2018pay} proposed DynamicConv and LightConv by replacing self-attention mechanism with a lightweight convolution.

\paragraph{Specialization of Attention Heads}
Since the publication of Transformer architecture~\citep{DBLP:conf/nips/VaswaniSPUJGKP17}, many researchers have been interested in analyzing how the attention mechanism works. ~\citet{DBLP:conf/acl/VoitaTMST19} systematically analyzed the attention heads in the encoder and categorized them into three functional subsets: positional, syntactic, and rare words. When dealing with dependency parsing, researchers also observed the same phenomenon that different heads could capture different syntactic functions~\citep{DBLP:journals/corr/abs-1911-12246,shen2020unsupervised}.

\begin{figure*}[tbp]
    \centering
    \includegraphics[width=0.78\linewidth]{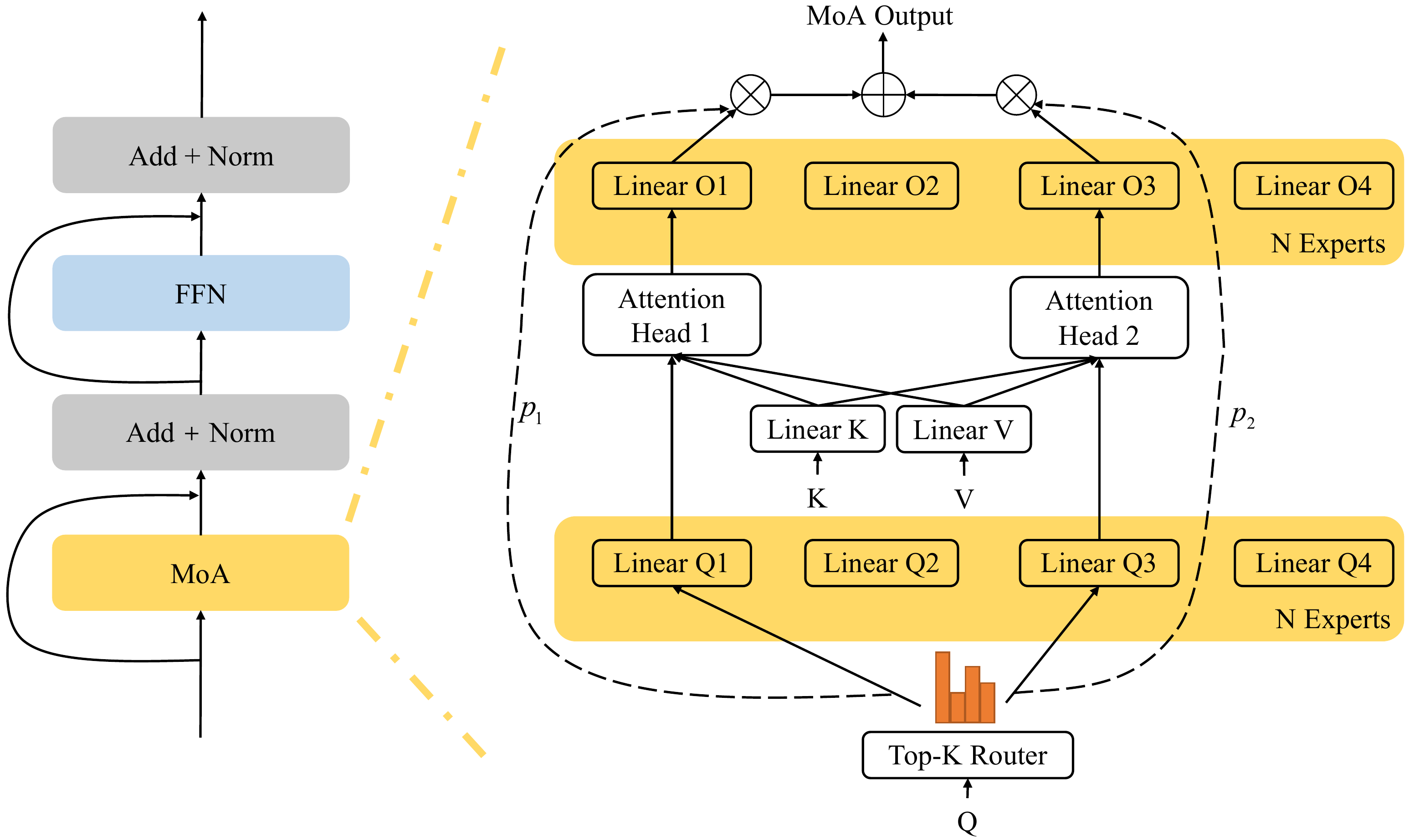}
    \caption{Mixture of Attention Heads (MoA) architecture. MoA contains two mixtures of experts. One is for query projection, the other is for output projection. These two mixture of experts select the same indices of experts. One routing network calculates the probabilities for each selected experts. The output of the MoA is the weighted sum of the outputs of each selected experts.}
    \label{fig:moa}
\end{figure*}

\section{Preliminaries}
\subsection{Mixture of Experts}
MoE~\citep{DBLP:conf/iclr/ShazeerMMDLHD17} contains a set of expert networks $E_1, E_2, \dots, E_N$ and a routing network $G$.
The output of the MoE is the weighted sum of the output of each expert. The routing network calculates the probability for each expert. Formally, the output of the MoE can be written as:
\begin{equation}
\label{eq:moe}
    y = \sum_{i=1}^N G(x)_i E_i(x)
\end{equation}

The routing network $G$ is a Noisy Top-k Routing network. Before the softmax function, they add Gaussian noise to the gated logits, see Equation~\ref{eq:gate}. Then, they keep only the top k values, setting the rest gate values to equal 0, see Equation~\ref{eq:noisytopk}.
\begin{equation}
\label{eq:noisytopk}
    G(x) = \operatorname{Softmax}(\operatorname{TopK}(H(x), k))
\end{equation}
\begin{align}
\label{eq:gate}
    H(x)_i =& (x\cdot W_g)_i + \sigma (0,1)\cdot\\ \notag
    &\operatorname{Softplus}((x\cdot W_{noise})_i)
\end{align}


\subsection{Multi-head Attention}
\citet{DBLP:conf/nips/VaswaniSPUJGKP17} proposed an encoder-decoder architecture Transformer, which contains the multi-head attention module. Different heads from the multi-head attention module attend to information from different representation subspaces, which learn the input from various perspectives.

Performing multi-head attention with $k$ heads, the $Q,K,V$ are linearly projected $k$ times with different, learned linear projections to subspaces. On each projected $Q$ and $K$, the attention scores are calculated, via Equation~\ref{eq:multiatt}. Values deriving from different heads are projected back to the model dimension size and summed up, with Equation~\ref{eq:attn}.
\begin{equation}
\label{eq:multiatt}
    W^{\rm att}_i = \text{Softmax} \left(\frac{QW_i^q(KW_i^{k})^T}{\sqrt{d_k}} \right)
\end{equation}
\begin{equation}
\label{eq:attn}
    y = \sum_{i=1}^k\left(W^{\rm att}_iVW_i^v\right)W_i^o
\end{equation}
where $W_i^q,W_i^k,W_i^v \in \mathbb{R}^{d_m\times d_h}$ and $W_i^o \in \mathbb{R}^{d_h\times d_m}$, $d_k$ is the dimension of the key $K$.

\section{Mixture of Attention Heads}
\label{sec:moa}
In this work, we propose a variant of multi-head attention for Transformer called Mixture of Attention Heads (MoA), illustrated in Figure~\ref{fig:moa}. 
MoA consists of two major components, the routing network $G$ and a group of $N$ attention experts $\left\{ E_1, ..., E_N \right\}$. 
Similar to standard multi-head self-attention, the input of MoA includes three sequences, query sequence $Q$, key sequence $K$, and value sequence $V$.
We note $q_t$ as the query vector at time step $t$.
For each $q_t$, the routing network $G$ selects a subset of $k$ experts $G(q_t) \subseteq \left\{ E_i \right\}$ based on $q_t$ and assign a weight $w_i$ to each selected expert.
Then, these selected experts take $q_t$, $K$, and $V$ as inputs and compute an output $E_i(q_t,K,V)$.
The output of the MoA is the weighted sum of the selected experts' outputs. 
Formally, the MoA output at time step $t$ can be written as:
\begin{equation}
    y_t = \sum_{i \in G(q_t)} w_{i,t}\cdot E_i(q_t,K,V)
\end{equation}

\subsection{Routing Network}
Similar to previous mixture-of-expert methods, the routing network assigns attention experts to input query.
In order to select $k$ experts for query $q_t$, we compute a routing probability $p_i$ for each expert $E_i$.
The routing probability is modeled with a linear layer $W_g$ and a softmax function: 
\begin{equation}
    p_{i,t} = \operatorname{Softmax}_i(q_t \cdot W_g)
\end{equation}
Based on the routing probability $p$, we select the top-$k$ attention experts among all $N$ attention experts with the largest probabilities. 
Formally, the routing network is defined as:
\begin{equation}
\label{eq:topk}
    G(Q)=\operatorname{TopK}(p_{i,t}, k)
\end{equation}
where $W_g\in \mathbb{R}^{d_m \times N}$, representing the routing matrix.
Then, we renormalize the routing probability of the selected experts to get normalized expert weights:
\begin{equation}
    w_{i,t} = \frac{p_i}{\operatorname{Detach} \left( \sum_{j \in G(q_t)} p_j \right)}
\end{equation}
where $\operatorname{Detach}(\cdot)$ is a function that stops the gradient backpropagation.
In other words, the denominator receives zero gradient during the training process.
We empirically find that this trick helps the routing network learn better routing probability.


\subsection{Attention Expert}
An attention expert contains four different projection matrices, $W^q$, $W^k$, $W^v$ and $W^o$. 
The attention calculation is similar to multi-head attention.
We first compute the attention weight for keys.
\begin{equation}
\label{eq:att}
    W^{\rm att}_{i,t} = \text{Softmax}\left(\frac{q_t W_i^{q}(KW^k)^T}{\sqrt{d_h}}\right)
\end{equation}
where $W_i^q \in \mathbb{R}^{d_m\times d_h}$ is the query projection matrix, $W^k \in \mathbb{R}^{d_m\times d_h}$ is the key projection matrix, $d_m$ is the hidden state size, $d_h$ is named as head dimension. 
We then compute the weighted sum of values:
\begin{equation}
    o_{i,t} = W^{\rm att}_{i,t} VW^v
\end{equation}
where $W^v \in \mathbb{R}^{d_m\times d_h}$ is the value projection matrix. 
Finally, the attention output is obtained by projecting $o_{i,t}$ back to the hidden state space:
\begin{equation}
\label{eq:moa}
    E_i \left( q_t,K,V \right) = o_{i,t} W_i^o
\end{equation}
where $W_i^o \in \mathbb{R}^{d_h\times d_m}$ is the output projection matrix.

In the multi-head attention, the projection matrices $W^q$, $W^k$, $W^v$, and $W^o$ are all different across attention heads. 
The MoA shares $W^k$ and $W^v$ across attention experts to reduce the computational complexity.
Attention experts are only differentiated by $W^q_i$ and $W^o_i$.
Thus, the expensive matrix projection of key sequence $KW^k$ and value sequence $VW^v$ can be pre-computed and shared for all attention experts.
Each expert only need to compute the vector projection of query $q_t W_i^q$ and output $o_{i,t} W_i^o$.
This design can significantly reduce the computational and space complexity while the number of experts is large.

\subsection{Training Losses}
Previous work~\citep{DBLP:conf/iclr/ShazeerMMDLHD17} has observed that the routing network tends to converge to a state where it always produces large weights for the same few experts, which indicates the insufficient utility of all the experts.
Following \citet{DBLP:conf/iclr/ShazeerMMDLHD17} and \citet{DBLP:journals/corr/abs-2101-03961}, we add an auxiliary loss to balance the loads of different experts.

Given $N$ experts and a sequence with $T$ queries $Q=\{q_1,q_2,\dots ,q_T\}$, the auxiliary loss $L_a$ can be computed as:
\begin{equation}
    L_a(Q) = N\cdot \sum_{i=1}^N f_i \cdot P_i
\end{equation}
where $f_i$ is the number of tokens attributed to the $i$-th expert, 
\begin{equation}
\label{eq:fi}
    f_i = \sum_{t=1}^T \delta_{i \in G(q_t)}
\end{equation}
where $\delta$ represents the Kronecker-Symbol.
$P_i$ is the sum of router probability allocated for the $i$-th expert,
\begin{equation}
\label{eq:pi}
    P_i = \sum_{t=1}^T p_{i,t}
\end{equation}
They are then normalized with norm 1 according to the expert column.
Mathematically, $f_i$ is indifferentiable while $P_i$ is. Thus, larger $f_i$ will result in a larger derivative. This penalizes the $P_i$ making larger $P_i$ smaller. What's more, $P_i$ is calculated by softmax. Thus smaller $P_i$ will become bigger.

\citet{DBLP:journals/corr/abs-2202-08906} introduced a router z-loss (Equation \ref{eq:zloss}) to penalize large logits into the gating network, which could stabilize the training and improve the performance.
\begin{equation}
\label{eq:zloss}
L_{z}(x)=\frac{1}{T} \sum_{j=1}^{T}\left(\log \sum_{t=1}^{N} e^{x_{i,t}}\right)^{2}
\end{equation}
where $x_{i,t}$ is the pre-softmax logit computed by router for $i$-th expert and input query $q_t$.
Each mixture of attention heads module has an auxiliary loss and a router z-loss. We sum them up together and added with a multiplicative coefficient $\alpha$ and $\beta$ respectively to the total model loss during training. Throughout this work, we use $\alpha = 0.01$ and $\beta=0.001$ to ensure the efficiency of the two added losses and not to disturb the primary cross-entropy model loss.
\begin{equation}
    L = L_{\text{model}} + \sum_{\forall \ \text{MoA module}} (\alpha L_a + \beta L_z)
\end{equation}

To validate the utility of these auxiliary losses, we conducted ablation tests and the results are shown in Appendix~\ref{sec:losses}.

\subsection{Computational Complexity and Number of Parameters}
On the one hand, given a sequence with $T$ tokens, the amount of computation required by an MoA layer that selects top-$k$ experts is
\begin{equation}
    C_{MoA} = kT^2d_h + 2(k+1)Td_hd_m
\end{equation}
where $k d_h$ is the sum of head dimension of selected experts.
It represents the maximum amount of information that can be collected by an MoA layer for a token.
On the other hand, the amount of computation required by a standard Multi-Head Attention (MHA) is 
\begin{equation}
    C_{MHA} = T^2d_m + 4Td_m^2
\end{equation}
where $d_m$ is the sum of head dimension.
If $kd_h\simeq d_m$, the computational complexity of MoA is smaller than that of MHA.
In other words, the MoA could collect more information for each token while maintaining a similar level of computational complexity as the MHA.

As for the number of parameters, given a Mixture of Attention Heads with $E$ attention experts, the number of parameters in MoA and MHA are:
\begin{equation}
    M_{MoA} = (2E+2)d_hd_m, \quad M_{MHA} = 4d_m^2 
\end{equation}
When $k = E$ and $Ed_h\simeq d_m$, the number of parameters in MoA is smaller than MHA.
In other words, MoA could collect more information for each token while maintaining a similar number of parameters as MHA.
More details of the calculation are in Appendix~\ref{sec:complex}.

The above discussion suggests that, from an information collection point of view, the MoA is more computational and parameter efficient than the standard MHA.
Our experimental results in Section~\ref{sec:results} also empirically support the hypothesis.
Additionally, the time complexity of MoA is decided by the number of attention heads $k$ and the attention head dimension $d_h$, not the model's total parameters.
One could arbitrarily increase the amount of parameters in MoA, without increasing its computational complexity.

\section{Experiments}
\label{sec:results}
\subsection{Machine Translation}

\paragraph{Dataset}
We train our Mixture of Attention model on WMT 2014 English-German and English-French datasets~\citep{DBLP:conf/wmt/BojarBFHKLMPPSS14}. 
Following the experimental settings used in \citet{liu2020very}, all sentences were encoded using byte-pair encoding~\citep{sennrich-etal-2016-neural}.
For both tasks, we use a joined dictionary and share all word embeddings of the encoder and the decoder.
For English-German, their shared vocabulary size is set to be 32k. 
For English-French, their shared vocabulary size is set to be 40k. 

\begin{table*}[htbp]
    \centering
    \small
    \begin{tabular}{rccccc}
    \toprule
        \multirow{2}{*}{Model} & 
        \multicolumn{2}{c}{WMT14 EnDe} & 
        \multicolumn{2}{c}{WMT14 EnFr} & 
        \multirow{2}{*}{MACs\footnotemark[3]}\\
        & \#Params & BLEU & \#Params & BLEU \\ 
        \midrule
        Transformer base~\citep{DBLP:conf/nips/VaswaniSPUJGKP17} & 65M & 27.3 & 62M & 38.1 & 604M \\
        Admin 6L-6L~\citep{liu2020very} & 61M & 27.7 & 67M & 41.5 & 604M \\
        MAE-7~\citep{DBLP:conf/acl/PengSLS20} & 63M & 28.4 & - & - & -\\ \midrule
        MoA Base ($8K8E128D$) & 65M & 28.4 & 69M & 42.5 & 628M \\ \midrule
        Transformer big~\citep{DBLP:conf/nips/VaswaniSPUJGKP17} & 213M & 28.4 & 210M & 41.8 & 2090M\\
        Transformer big~\citep{ott-etal-2018-scaling} & 210M & 29.3 & 222M & 43.2 & 2090M \\
        LightConv~\citep{wu2018pay} & 202M & 28.9 & - & 43.1 & 1750M\footnotemark[4]\\
        DynamicConv~\citep{wu2018pay} & 213M & 29.7 & - & 43.2 & 1790M\footnotemark[4] \\
        Admin 18L-18L~\citep{liu2020very} & 151M & 29.0 & - & - & 1490M \\
        Admin 60L-12L~\citep{liu2020very} & 256M & 30.1 & 262M & 43.8 & 2550M \\
        \midrule
        MoA Big ($16K32E256D$) & 200M & 29.4 & 204M & 43.7 & 1220M \\ \bottomrule
    \end{tabular}
    \caption{BLEU score on WMT14 translation datasets. 
    MACs (Multiply–Accumulate Operations)\footnotemark[3] measures the computational complexity of each model.
    For different models, their MACs are computed on a source sentence of length $T_{src}=10$ and a target sentence of length $T_{tgt}=10$.
    }
    \label{tab:bleu}
\end{table*}

\paragraph{Training and Evaluation Details}
We use the Adam Optimizer~\citep{DBLP:journals/corr/KingmaB14} with a learning rate of $7e^{-4}$ and the inverse square root learning rate scheduler.
During training, we employed label smoothing~\citep{DBLP:conf/cvpr/SzegedyVISW16} of value 0.1.
More training details can be found in Appendix~\ref{app:training}.

For the evaluation, we average the last 10 epochs' checkpoints.
We list BLEU score~\citep{DBLP:conf/acl/PapineniRWZ02} computed with \textsc{multi-bleu.perl}, and apply the compound split post-processing\footnote{\url{https://github.com/tensorflow/tensor2tensor/blob/master/tensor2tensor/utils/get_ende_bleu.sh}} introduced in \citet{DBLP:conf/nips/VaswaniSPUJGKP17}. 
We use MACs (Multiply–Accumulate Operations)\footnotemark[3] to evaluate the computational complexity of different models on a fixed input. 
Details of the MACs calculation are in Appendix~\ref{app:macs}.

\footnotetext[3]{We adopt the open-source tool \textsc{ptflops} (\url{https://github.com/sovrasov/flops-counter.pytorch}) to calculate the MACs.}
\footnotetext[4]{These MACs values are underestimated.
Because the \textsc{ptflops} does not support the customized convolution layers in DynamicConv and LightConv. }

\paragraph{Baselines}
We compare with several strong baselines: Transformer base and big~\citep{DBLP:conf/nips/VaswaniSPUJGKP17}, Transformer big~\citep{ott-etal-2018-scaling} with reduced precision and large batch training, DynamicConv~\citep{wu2018pay} by replacing self-attention mechanism with a lightweight convolution, MAE-7 with reallocation of attention heads proposed by ~\citet{DBLP:conf/acl/PengSLS20}, Admin~\citep{DBLP:conf/emnlp/LiuLGCH20} which deepens the Transformer architecture.

For our model, three parameters are used to differentiate its variants, one is number of the activated attention heads ($K$) per token, one is total number of the experts ($E$), another is the attention expert dimension ($D$). 
For example, our MoA base model is noted as $8K8E128D$, because it has 8 attention experts, 128 dimension per expert, and all 8 experts are activated for each token.
Our MoA big model is $16K32E256D$ as it has 32 attention experts and sparsely activates the top 16 experts for each token.



\paragraph{Results}
The results on the test set of WMT14 EnDe and WMT14 EnFr datasets are shown in Table~\ref{tab:bleu}. The table is split into 2 parts, the upper part is for base models and the lower part is for large models.
On all datasets, MoA base outperforms Transformer base and Admin 6L-6L by at least 0.6 BLEU. 
On the WMT14 EnFr dataset, MoA base also outperforms Transformer big. 
On the WMT14 EnDe dataset, MoA base reaches comparable results with the Mixture of Attention Experts model (MAE-7), which is the state-of-the-art performance for base-level models. MACs of MAE-7 and our model are comparable in the setting of 8 attention heads.
While both models leverage the idea of weighting the attention heads, MoA is easier to implement and does not require the complicated block coordinate descent training steps.
Compared to standard multi-head self-attention, the routing mechanism pays more attention to the more informative attention heads for each token, thus enabling the MoA base model to achieve better computation and parameter efficiency.


In the big-scale setting, MoA big consistently outperforms standard transformer big models, despite requiring significantly less computation.
Compared to the models with more parameters, MoA is still very competitive. 
Only Admin 60L-12L outperforms MoA big on both datasets. 
However, the model has more parameters and requires about two times of MACs.
The MACs of MoA big is 1220M, which is the lowest amount among big-scale models.
This result shows that our proposed method could easily scale up to a large amount of parameters and achieve good results without substantially burdening the computation system.

\begin{table}[t]
    \centering
    \small
    \begin{tabular}{ccccc}
        \toprule
        \multicolumn{2}{c}{Model} & \#Params & PPL & MACs \\ 
        \midrule
        \multicolumn{2}{c}{Transformer} & 51.34M  &  4.95 & 6.55G \\ 
        \midrule
        \multirow{4}{*}{MoA} & $8K8E128D$ & 52.45M & 4.82 &  7.27G \\
         & $8K8E256D$ & 61.89M & 4.64 & 8.97G \\
         & $8K16E256D$ & 78.75M & 4.48 & 8.97G \\
         & $8K32E256D$ & 112.47M & 4.25 & 8.98G \\
         & $8K64E256D$ & 179.91M & 4.21 & 8.98G\\
        \bottomrule
    \end{tabular}
    \caption{Perplexity on wikitext-103 corpus test data for masked language modeling. 
        MACs are computed on a input sequence of length $T=128$.}
    \label{tab:ppl}
\end{table}

\begin{table*}[htbp]
    \small
    \centering
    \begin{tabular}{rccccccc}
    \toprule
        MoA Model & $K$ & $E$ & $D$ & \#Params & PPL(Valid) & BLEU(Test) & MACs \\ 
        \midrule
        Base & 8 & 8 & 128 & 65M & 4.68 & 28.4 & 628M \\ 
        \midrule
        (A) & 8 & 8 & 256 & 87M & 4.51 & 28.7 & 841M \\ 
        (B) & 8 & 16 & 256 & 125M & 4.45 & 28.4 & 841M \\ 
        (C) & 8 & 32& 64 & 83M & 4.79 & 27.9 & 524M \\ 
        (D) & 8 & 32& 128 & 123M & 4.55 & 28.4 & 631M \\ 
        (E) & 8 & 32& 256 & 200M & 4.44 & 28.8 & 841M \\ 
        (F) & 4 & 32& 256 & 200M & 4.65 & 27.5 & 654M \\ 
        \midrule
        Big & 16 & 32& 256 & 200M & 4.35 & 29.4 & 1220M \\ 
        \bottomrule
    \end{tabular}
    \caption{BLEU score of different MoA models on the WMT14 EnDe Dataset.
    }
    \label{tab:moaarch}
\end{table*}

\subsection{Masked Language Modeling}

Masked Language Modeling is the standard training objective for many Pretrained Language Models (PLMs), including BERT~\citep{DBLP:conf/naacl/DevlinCLT19} and RoBERTa~\citep{DBLP:journals/corr/abs-1907-11692}.
The task replaces a random sample of tokens in the input sequence with a special token \texttt{[MASK]}. 
The training objective is a cross-entropy loss on predicting the masked tokens.
To better mimic the procedure of training PLMs, we adopt the setting introduced in RoBERTa~\citep{DBLP:journals/corr/abs-1907-11692} to conduct the masked language modeling experiment. 

\paragraph{Dataset}
We conducted the masked language modeling on the wikitext-103 dataset~\citep{merity2016pointer}. 
The corpus includes over 100 million tokens collected from verified Good and Featured articles on English Wikipedia.
Following the settings in \citet{merity2016pointer}, the training/validation/test set has 103M/218K/246K words.
The corpus is tokenized with the 50K subword vocabulary used in RoBERTa and initially introduced in GPT~\citep{radford2019language}. 

\paragraph{Settings}
Then we train the model with the dynamic masking strategy and full-sentences input format.
To avoid overfitting on the training corpus, we adopt a medium-size RoBERTa model as the base model, with 512-dim word embedding, 2048-dim feed-forward network, 8 heads, and 8 layers. 
Training details can be found in Appendix~\ref{app:training}.
The perplexity is used as the evaluation metric.

\paragraph{Results}
Table~\ref{tab:ppl} shows the perplexity on WikiText-103 test data.
While using a similar amount of parameters, MoA outperforms the standard transformer model by 0.13 perplexity. 
Furthermore, the performance simultaneously improves with the increase of number of experts $E$ and head size $D$, while the number of selected heads $K$ and the computational complexity remains the same.
The observation shows that our model has the ability of increasing the model's performance while maintaining the computation complexity.

\subsection{Model Analysis}

\paragraph{MoA parameter influence}
We study the influence of three parameters, $K$, $E$, and $D$, on the WMT14 En-De Dataset. The results are shown in Table~\ref{tab:moaarch}. 
For the expert dimension $D$, we control $K=8$ and $E=32$, vary the expert dimension $D$ with 64, 128, and 256. As the expert  dimension size $D$ increases (rows C, D, E in Table~\ref{tab:moaarch}), the PPL on the  validation set and the BLEU score on the test set both improve. This amelioration is due to the increase in parameters. With a larger expert dimension size, each expert has more parameters, and the computational costs increase. We believe the increase in computational cost is acceptable. As in Table~\ref{tab:bleu}, Transformer big model has a MACs of 2090M, reaching BLEU of 28.4. However, by enlarging hidden size of expert, we could get BLEU of 28.8 while MACs at 841M (<<2090M).

For the number of attention experts $E$, we control $K=8$ and $D=256$, select three different values of $E$, 8, 16, and 32. When adding the number of experts, the PPL on the valid set goes down, indicating our model's continuous scaling up ability. The BLEU score on the test set does not change with that of PPL, and this may be because the training objective is not directly linked to the BLEU score calculation. 
However, we still observe that 32 experts can achieve better BLEU than the other two settings. 
As the number of selected attention heads $K$ remains unchanged, the MACs for these three settings are the same. 
Thus, MoA allows us to improve the model ability by adding more parameters without changing the computational complexity.

For the number of selected attention heads $K$, we test three numbers of selected attention heads $K$, 4, 8, and 16, freezing $E=32$ and $D=256$. With the increase in the number of selected attention heads, we observe that the PPL on the valid set decreases and the BLEU score on the test set goes up. Since the number of attention experts remains the same, the model's total parameters stay at 200M. 
This result shows the trade-off between computation efficiency and performance. The model needs more computations for better performance as the MACs vary from 654M to 1220M.

\paragraph{MoA Expert loads}
\begin{figure}
    \centering
    \includegraphics[width=0.9\columnwidth]{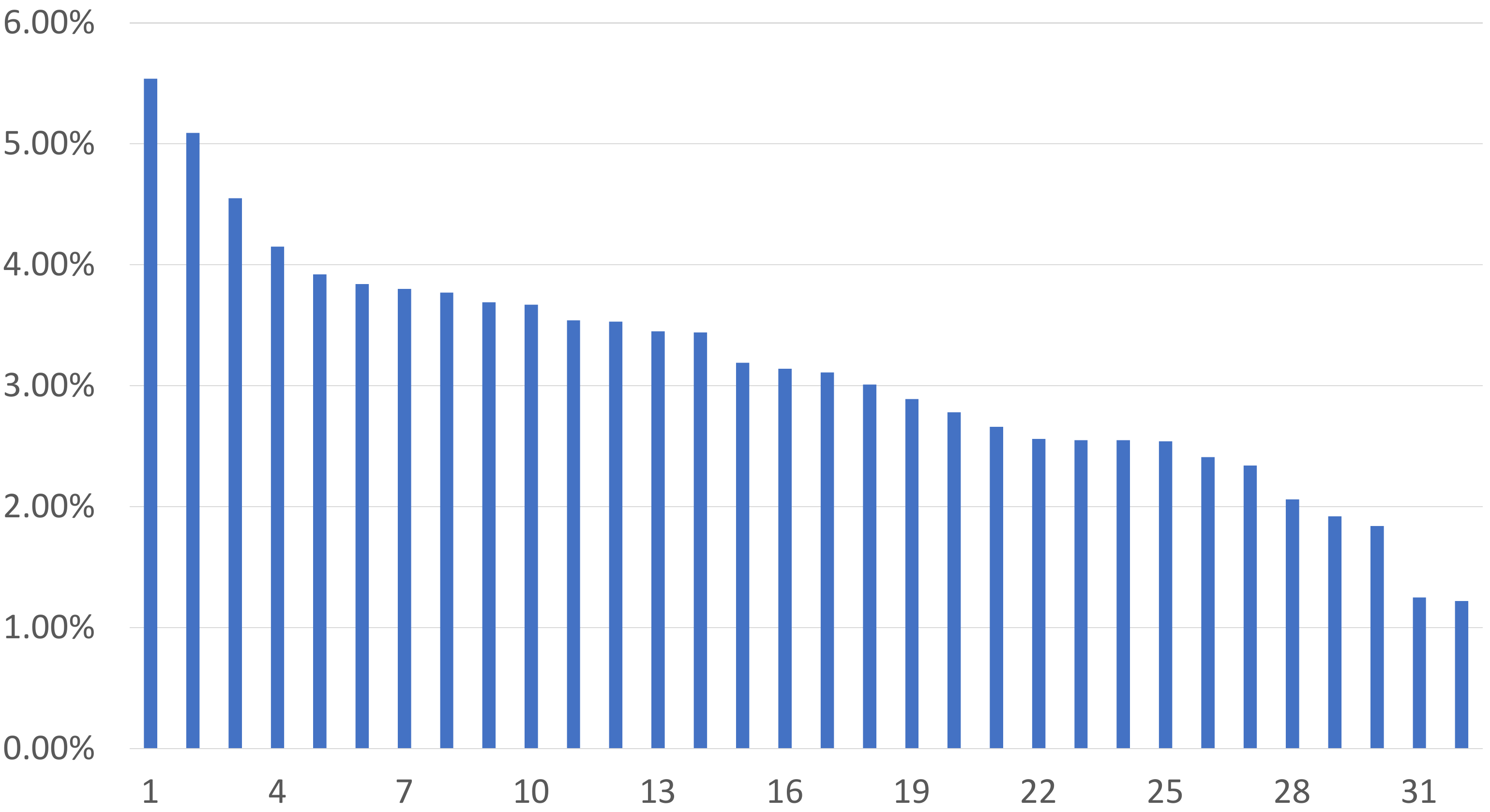}
    \caption{Experts' load percentages for encoder layer 4. 
    Experts are indexed by their order of percentages.
    }
    \label{fig:percent4}
\end{figure}

Load balancing is a long-standing problem of MoE models~\citep{DBLP:journals/corr/abs-2101-03961}.
Figure~\ref{fig:percent4} shows the experts' load of the fourth layer of the MoA big model.
It plots the percentage of each expert used in the development set of WMT14 EN-DE.
For each input token and MoA big, 16 attention heads are selected among 32 attention experts. 
This figure shows a relatively based load, where the most used expert is selected by 5\% of the tokens and the least used expert is selected by 1\%, and most experts' loads are between 2-4\%.
This observation suggests that the input tokens are attributed equally among attention experts. 
Experts are assigned to different roles with substantially different groups of input tokens.
The load for every expert of different layers is shown in Appendix~\ref{sec:percent}. 

\paragraph{MoA Interpretability: Specialize the Experts}


\begin{table}[t]
    \centering
    \small
    \begin{tabular}{ccccccc} \toprule
        Expert 5 &Expert 29 & Expert 10 \\ \midrule
        \textbf{Tech.} & \textbf{adv.} & \textbf{Location}  \\ \midrule
        DSL & likely & Costasur \\
        JPEG &tastefully & Kaliningrad\\
        Module & environmentally & Freiburg  \\
        DSLR &  heavily & Brava  \\
        screen & certainly  & Jesolo \\ \midrule
        Expert 7 & Expert 24 & Expert 23 \\ \midrule
        \textbf{Computer} & \textbf{Reaction} & \textbf{Name} \\ \midrule
        Outlook & supportive & Marx\\
         Excel & misunderstanding & Jarzembowski \\
        IO & advocating & Reding\\
        emails & confirming & Donald\\
        monitors & excitement &Socrates \\
        \bottomrule
    \end{tabular}
    \caption{Indicative tokens of each expert for the first encoder layer of MoA}
    \label{tab:indicative}
\end{table}

We study in this section whether the different experts possess different ``expertises''.
We try to find the most likely tokens to co-occur with each expert. 
We compute the pointwise mutual information (PMI;~\citealt{DBLP:journals/coling/ChurchH90}) between tokens and experts:
\begin{equation*}
    \operatorname{PMI}({\rm token}_i, {\rm expert}_j) = \frac{p({\rm token}_i, {\rm expert}_j)}{p({\rm token}_i)\cdot p({\rm expert}_j)} \text{.}
\end{equation*}
For each expert, the bigger the PMI, the more relevant the token with this expert. 
Table~\ref{tab:indicative} lists the most indicative tokens of each expert for the first encoder layer of $16K32E512D$. 
Many experts are associated with nouns in the same topic, e.g., Location, Name, Tech, etc. 
We also found that some other experts are associated with adjectives and adverbs. 
For example, Expert 29 is related to adverbs, and Expert 24 is connected to people's reactions, where some tokens are adjectives. 
We also study the relation between expert and input tokens for other layers of the encoder, but it is hard to find clear patterns in other layers.

\section{Conclusion}
This work introduces the Mixture of Attention Heads (MoA). 
MoA contains a set of attention experts and a routing network. 
Given an input, the MoA attributes a probability to each expert by the routing network and selects the top-K experts.
The output of MoA is a weighted sum of the selected attention experts. 
The weighting mechanism allows different tokens to focus on different experts, thus improving the model's parameter and computation efficiency.
Experimental results show that a base-scale MoA model could achieve comparable or better performance than a transformer big model.
Furthermore, MoA could improve its performance by adding more attention experts while maintaining a relatively small computational complexity. 
In this way, MoA can achieve comparable performance with deeper and computationally more expensive models.
The interpretability analysis shows that different attention experts tend to specialize in a specific type of input tokens.

\section*{Limitations}
In this work, we scale up MoA to at most 64 experts. However, regarding the works combining mixture of experts with FFN layer, they could expand the expert number to thousands. In the future, we will explore the limit of the scale up ability of MoA.

Our implementation of MoA is not optimistic. Our code could not fully explore the parallel computing capability of GPUs. Our current implementation spends some extra time on memory copy operations. 
Although the computational complexity (MACs) of MoA is relatively low compared to other baselines, the running time of our implementation is not optimal. In the future, if we could optimize the implementation at the cuda kernel level to remove the memory copy ops, we expect at least half the wall-clock time. This will make an MoA block as fast as a standard attention block. 

Similar to Transformer architecture, MoA needs a careful hyperparameter search to reach satisfying results.

\section*{Acknowledgments}
This work is supported in part by the State Key Laboratory of the Software Development Environment of China under Grant SKLSDE-2021ZX-16.

\bibliography{anthology}
\bibliographystyle{acl_natbib}

\newpage
\appendix
\onecolumn
\section{Experts' load percentages}
\label{sec:percent}

\begin{figure*}[htbp]
    \centering
    \includegraphics[width=0.8\linewidth]{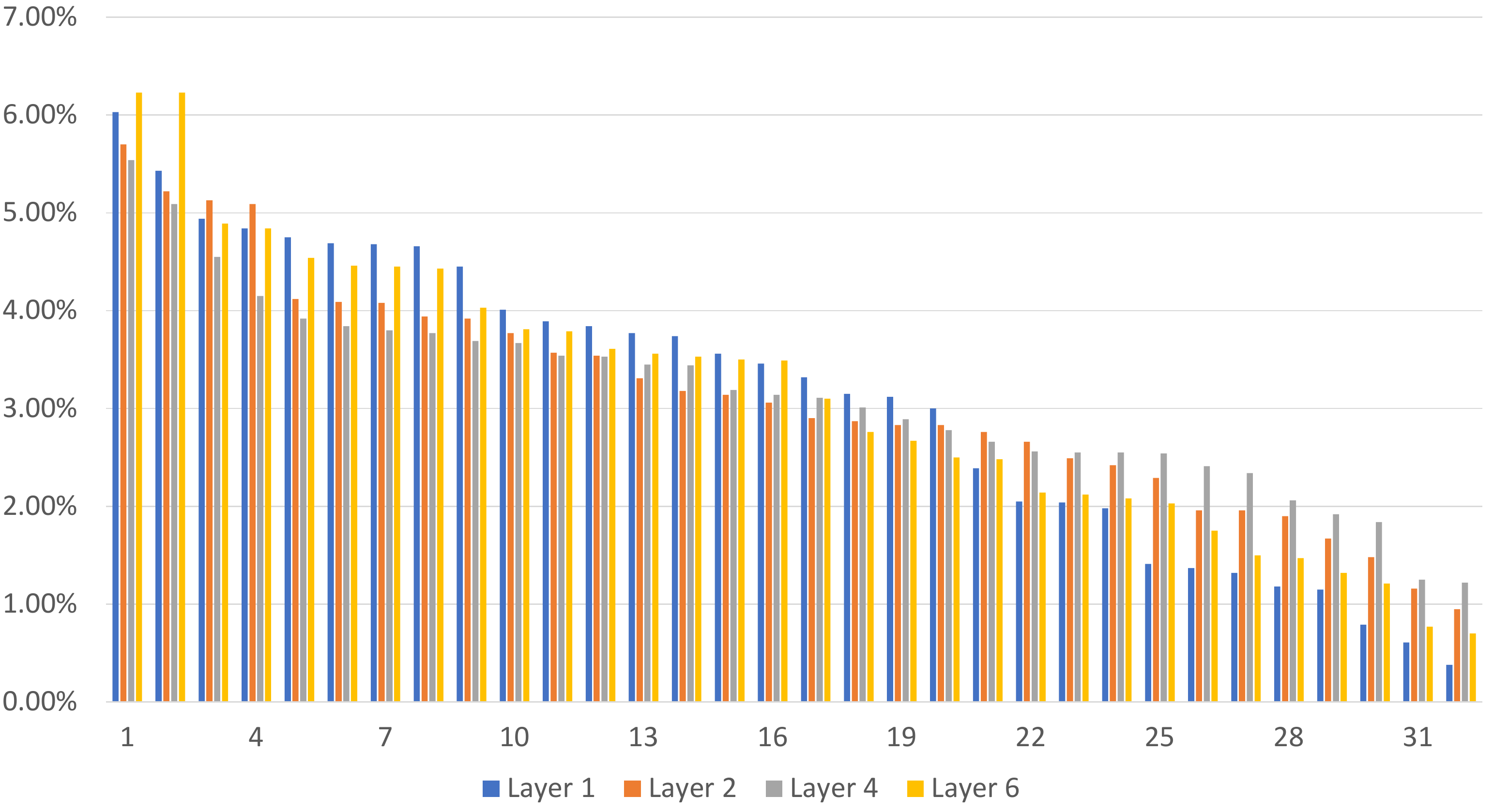}
    \caption{Experts' load percentages for different encoder layers}
    \label{fig:percent}
\end{figure*}

We compare the attribute distribution of tokens on different experts for different encoder layers of $16K32E512D$. The results are shown in Figure~\ref{fig:percent}. The load percentages for each expert in different layers are relatively balanced.

\section{Computational Complexity Proof}
\label{sec:complex}
Given a Mixture of Attention Heads with $E$ attention experts, a MoA layer has $(2E+2)d_{\rm head}d_{\rm model}$ parameters. A multi-head attention layer has $4d_{\rm model}^2$ parameters.
To compare these two complexities, we conduct the fraction of them.
\begin{align*}
    &\frac{2(E+1)d_{\rm head}d_{\rm model}}{4d_{\rm model}^2} \\
    =&\frac{(E+1)}{2K} \frac{Ed_{\rm head}}{d_{\rm model}}
\end{align*}
We note $q=\frac{Ed_{\rm head}}{d_{\rm model}}$, then we get
\begin{align*}
    \left(\frac{1}{2} + \frac{1}{2E}\right)q
\end{align*}
When $Ed_{\rm head}\simeq d_{\rm model}$, we have $q\simeq 1$, and
\begin{align*}
    \frac{1}{2} + \frac{1}{2E}
\end{align*}
which is a hyperbolic-like curve, with value equals to 1 when $E = 1$.
Therefore, if $E>1$, we have the proportion between the parameters of MoA and that of multi-head attention inferior to 1. Thus, the MoA layer contains fewer parameters than multi-head attention layer.



\section{Traning Details}
\label{app:training}

\begin{table*}[htbp]
    \centering
    \small
    \begin{tabular}{ccccccccccccc}
        \toprule
        Dataset & Model & Emb Size & FFD Size & Encoder Layers & Decoder Layers \\ \midrule
        \multirow{2}{*}{WMT14} & MoA base & 512 & 2048 & 6 & 6  \\
        & MoA big & 512 & 2048 & 6 & 6 \\
        Wikitext-103 & MoA & 512 & 2048 & 8 & -  \\
        \bottomrule
    \end{tabular}
    \caption{Hyperparameters for different models. 
    }
    \label{tab:model_hyperparameters}
\end{table*}

\begin{table*}[htbp]
    \centering
    \small
    \begin{tabular}{ccccccccccccc}
        \toprule
        Dataset & Model & BSZ & LR & warmup & Dropout & DropATT & DropFFD & Epochs \\ \midrule
        \multirow{2}{*}{WMT14 EN-DE} 
        & MoA base & 8092 $\times$ 32 & 7e-4 & 4000 & 0.2 & 0.2 & 0.1 & 100 \\
        & MoA big & 4096 $\times$ 64 & 7e-4 & 4000 & 0.2 & 0.2 & 0.1 & 100 \\
        \multirow{2}{*}{WMT14 EN-FR} 
        & MoA base  & 8092 $\times$ 32 & 7e-4 & 8000 & 0.1 & 0 & 0.1 & 50 \\
        & MoA big & 4096 $\times$ 64 & 7e-4 & 8000 & 0.1 & 0.1 & 0.1 & 100 \\
        Wikitext-103 & MoA & 16384 $\times$ 32 & 6e-4 & 2000 & 0.1 & 0 & 0 & 60 \\
        \bottomrule
    \end{tabular}
    \caption{Training Hyperparameters for different models. 
    The BSZ represent the maximum number of tokens in each batch.
    }
    \label{tab:hyperparameters}
\end{table*}
All of our models are trained on 32 V100 GPUs.
We use the Adam Optimizer~\citep{DBLP:journals/corr/KingmaB14} with $\beta_1= 0.9$, $\beta_2= 0.98$ and $\epsilon = 1e-9$. 
We use a inverse square root learning rate scheduler for the translation tasks and a linear scheduler for the masked language model task.
During training, we employed label smoothing~\citep{DBLP:conf/cvpr/SzegedyVISW16} of value 0.1. 
More training hyperparameters can be found in Table~\ref{tab:hyperparameters}.

\section{Utility of different auxiliary losses}
\label{sec:losses}
We adopted two different auxiliary losses to balance the experts' loads, one is $L_a$ proposed by~\citet{DBLP:journals/corr/abs-2101-03961}, the other is $L_z$ proposed by~\citet{DBLP:journals/corr/abs-2202-08906}. To validate the utility of these two auxiliary losses, we conducted several ablation tests. The results are shown in Table~\ref{tab:losses}. With different combinations of auxiliary losses and different coefficients, we found that 0.01$L_a$ + 0.001$L_z$ achieved the best BLEU score on WMT14 EnDe test set.

\begin{table}[htbp]
    \centering
    \small
    \begin{tabular}{cccccc} \toprule
        MoA & 0.01$L_a$ & 0.01$L_z$ & 0.001$L_z$ & 0.01$L_a$+0.001$L_z$ & 0.01$L_a$+0.01$L_z$ \\ \midrule
        $8K8E128D$ & 28.95 & 28.73 & 28.78 & 28.94 & 28.73\\
        $8K16E128D$ & 28.53 & 28.68 & 28.61 & 28.77 & 28.62\\
        $8K32E128D$ & 28.45 & 28.31 & 28.38 & 28.32 & 28.4\\ \bottomrule
    \end{tabular}
    \caption{Ablation test for different auxiliary losses}
    \label{tab:losses}
\end{table}

\section{MACs calculation}
\label{app:macs}
\textsc{ptflops} launches a given model on a random tensor (with pre-defined input shapes) and estimates amount of computations (multiply-add operations) during inference. We need to define the shapes of inputs when using \textsc{ptflops} to calculate MACs. For translation models, we set encoder sequence length and decoder sequence length to be 10. We set the batch size to be 1. For language modeling models, we set the sequence length to be 128 and the batch size to be 1. With the pre-defined input shapes, \textsc{ptflops} will conduct the forward process of the given model and feed back the MACs.


\end{document}